\documentclass[journal]{IEEEtai}

\usepackage[colorlinks,urlcolor=blue,linkcolor=blue,citecolor=blue]{hyperref}

\usepackage{color,array}

\usepackage{graphicx}

\usepackage{url}            % simple URL typesetting
\usepackage{booktabs}       % professional-quality tables
\usepackage{amsfonts}       % blackboard math symbols
\usepackage{nicefrac}       % compact symbols for 1/2, etc.
\usepackage{microtype}      % microtypography
\usepackage{xcolor}         % colors
\usepackage{float}          % table position
\usepackage{adjustbox}
\usepackage{subcaption}
\usepackage{colortbl}
\usepackage{threeparttable}
\usepackage{bbding}
\usepackage{verbatim}
\usepackage{bm}
\usepackage{multirow}
\usepackage{tabularx}
\usepackage{pifont}
\usepackage{amsmath}
\usepackage{algpseudocode}
\begin{document}

\title{Representation Finetuning for Continual Learning} 
\author{Haihua Luo*, Xuming Ran*, Tommi Kärkkäinen, Huiyan Xue, Zhonghua Chen, Qi Xu, Fengyu Cong
% \author{First A. Author, \IEEEmembership{Fellow, IEEE}, Second B. Author, and Third C. Author, Jr., \IEEEmembership{Member, IEEE}

% \thanks{This paragraph of the first footnote will contain the date on which you submitted your paper for review. It will also contain support information, including sponsor and financial support acknowledgment. For example, ``This work was supported in part by the U.S. Department of Commerce under Grant BS123456.'' }
% \thanks{The next few paragraphs should contain the authors' current affiliations, including current address and e-mail. For example, F. A. Author is with the National Institute of Standards and Technology, Boulder, CO 80305 USA (e-mail: author@boulder.nist.gov).}
% \thanks{S. B. Author, Jr., was with Rice University, Houston, TX 77005 USA. He is now with the Department of Physics, Colorado State University, Fort Collins, CO 80523 USA (e-mail: author@lamar.colostate.edu).}
% \thanks{T. C. Author is with the Electrical Engineering Department, University of Colorado, Boulder, CO 80309 USA, on leave from the National Research Institute for Metals, Tsukuba, Japan (e-mail: author@nrim.go.jp).}
% \thanks{This paragraph will include the Associate Editor who handled your paper.}}

\thanks{H. Luo, T. Kärkkäinen and Z. Chen are with the Faculty of Information Technology, University of Jyväskylä, Jyväskylä, Finland (e-mail: {haihua.h.luo, tommi.karkkainen, zhonghua.x.chen}@jyu.fi).}
\thanks{X. Ran is with National University of Singapore, Singapore, Singapore (e-mail: {ranxuming}@gmail.com).}
\thanks{H. Xue, Q. Xu and F. Cong are with School of Computer Science and Technology, Dalian University of Technology, Dalian, China (e-mail: {xuebb, xuqi, cong}@dlut.edu.cn).}
\thanks{* Equal contribution}
\thanks{Correspondence to: Qi Xu ({xuqi}@dlut.edu.cn)}
}

% \markboth{Journal of IEEE Transactions on Artificial Intelligence, Vol. 00, No. 0, Month 2020}
% {First A. Author \MakeLowercase{\textit{et al.}}: Bare Demo of IEEEtai.cls for IEEE Journals of IEEE Transactions on Artificial Intelligence}

\maketitle

\begin{abstract}
The world is inherently dynamic, and continual learning aims to enable models to adapt to ever-evolving data streams. While pre-trained models have shown powerful performance in continual learning, they still require finetuning to adapt effectively to downstream tasks. However, prevailing Parameter-Efficient Fine-Tuning (PEFT) methods operate through empirical, black-box optimization at the weight level. These approaches lack explicit control over representation drift, leading to sensitivity to domain shifts and catastrophic forgetting in continual learning scenarios. In this work, we introduce Continual Representation Learning (CoRe), a novel framework that  for the first time shifts the finetuning paradigm from weight space to representation space. Unlike conventional methods, CoRe performs task-specific interventions within a low-rank linear subspace of hidden representations, adopting a learning process with explicit objectives, which ensures stability for past tasks while maintaining plasticity for new ones. By constraining updates to a low-rank subspace, CoRe achieves exceptional parameter efficiency. Extensive experiments across multiple continual learning benchmarks demonstrate that CoRe not only preserves parameter efficiency but also significantly outperforms existing state-of-the-art methods. Our work introduces representation finetuning as a new, more effective and interpretable paradigm for continual learning.

\end{abstract}

\begin{IEEEImpStatement}
Continual learning aims to enable neural network models adapted to evolving data streams, but traditional finetuning methods often suffer from catastrophic forgetting and parameter inefficiency. Our work introduces Continual Representation Learning(CoRe), a novel framework that applies representation finetuning to continual learning for the first time. By performing interventions in low-rank subspaces of hidden representations, CoRe achieves superior parameter efficiency and mitigates forgetting without relying on black-box optimization. Extensive experiments demonstrate that CoRe consistently outperforms existing methods across multiple benchmarks. This approach enhances the adaptability of pre-trained models in dynamic environments, making it suitable for real-world applications such as autonomous systems, robotics, and personalized artificial intelligence assistants, where efficient, lifelong learning is critical.
\end{IEEEImpStatement}


\section{Introduction}
\label{sec:intro}

The world is dynamically changing, however, machine learning models are usually trained under the assumption that the training and test data come from the same stationary distribution. When exposed to non-stationary data streams, such models often suffer from Catastrophic Forgetting\cite{goodfellow2013empirical}, losing previously acquired knowledge when learning new tasks. This challenge has motivated extensive research in continual learning\cite{de2021continual,masana2022class}, whose goal is to enable models to incrementally acquire new knowledge without sacrificing performance on prior tasks.

Recently, continual learning methods with pre-trained models have attracted increasing attention\cite{wang2022learning, wang2022dualprompt, yu2024boosting}. Pre-trained models, such as ViT\cite{dosovitskiy2020image}, are extensively trained on large-scale datasets and demonstrate powerful feature extraction capabilities. However, a domain gap often exists between pretraining datasets and downstream datasets\cite{zhou2025revisiting}. As a result, pre-trained models typically require finetuning to adapt effectively to new tasks. Traditional finetuning methods, such as full-finetuning and parameter-efficient fine-tuning (PEFT)\cite{houlsby2019parameter,jia2022visual}, primarily focus on updating model parameters. Whereas effective, these approaches predominantly operate through implicit, black-box optimization in the weight space, lacking interpretability during the learning process and making it difficult to directly explain the role of the updated parameters. Besides, their learning process is also largely empirical and objectives are unclear. This paradigm lacks of explicit control over representation drift makes the model susceptible to interference between sequential tasks, often leading to suboptimal stability-plasticity trade-off and sensitivity to domain shifts and class imbalance. Moreover, they still require a relatively large number of parameters and the parameter efficiency needs to be further improved.

Unlike weight-based finetuning, the recently proposed representation finetuning (ReFT)\cite{wu2024reft} directly intervenes on a model’s hidden representations, often within a low-rank linear subspace, providing a more flexible and efficient way to adapt models. By enabling task-specific interventions without introducing excessive parameter overhead, ReFT has achieved competitive results across various domains\cite{liu2025unlocking, yin2024lofit, osial2025parameter, huang2025enhancing}, particularly in large language models. Nevertheless, despite its effectiveness, ReFT has not yet been explored in the context of continual learning, where controlling representation drift is especially critical.

In this work, we propose the \textbf{Co}ntinual \textbf{Re}presentation Learning(\textbf{CoRe}) framework, the first approach to integrate representation finetuning with continual learning. Unlike traditional weight-based PEFT methods, CoRe constructs task-specific low-rank intervention subspaces within critical representation layers of the model. By leveraging explicit optimization goals to guide the evolution of representations, CoRe achieves efficient adaptation to new tasks while effectively mitigating catastrophic forgetting. Adopting low-rank subspace interventions, CoRe attains superior parameter efficiency and demonstrates outstanding performance across multiple benchmarks.

Our contributions can be summarized as follows:
\begin{itemize}
    \item We propose CoRe, the first framework to integrate representation finetuning into continual learning, bridging the gap between representation-level interventions and incremental task adaptation.
    \item CoRe performs task-specific interventions in the low-rank linear subspace of hidden representations with explicit objectives, ensuring parameter efficiency and improving adaptability.
    \item Experiments on extensive continual learning benchmarks show that CoRe consistently outperforms existing parameter-efficient fine-tuning methods, achieving state-of-the-art results and maintaining efficiency.
\end{itemize}

\section{Related Work}
\label{sec:related_work}

\subsection{Continual Learning}
\label{subsec:continual_learning}
Traditional continual learning approaches can generally be categorized into three groups: regularization-based, architecture-based, and replay-based methods. Regularization-based approaches\cite{aljundi2018memory, serra2018overcoming, li2017learning} preserve knowledge from previous tasks by imposing constraints on parameter updates; however, such constraints may also hinder the model’s ability to acquire knowledge from new tasks. Architecture-based approaches\cite{mallya2018packnet, serra2018overcoming, wang2020learn} address this limitation by dynamically modifying the model structure to accommodate new tasks, but this often results in increased memory consumption. Replay-based approaches\cite{rebuffi2017icarl, buzzega2020dark, cha2021co2l}, on the other hand, maintain a memory buffer that stores data or knowledge from past tasks, which can be replayed during the training of new tasks to mitigate forgetting. Nevertheless, these methods face challenges such as continually growing memory requirements and potential privacy concerns. Recently, continual learning methods built upon pretrained models have attracted significant attention\cite{wang2022learning, wang2022dualprompt, yu2024boosting}. Pretrained models, such as ViT\cite{dosovitskiy2020image} and CLIP\cite{radford2021learning}, are extensively trained on large-scale datasets and exhibit strong feature extraction capabilities. However, a domain gap often exists between the pretraining datasets and the downstream datasets in continual learning scenarios\cite{zhou2025revisiting}. As a result, pretrained models typically require finetuning to adapt effectively to downstream tasks.

\subsection{Representation Finetuning}
\label{sbusec:represen_finetune}
Representation finetuning methods intervene in models by modifying the semantic representations of inputs using counterfactual information. Representation finetuning (ReFT)\cite{wu2024reft} first introduce representation finetuning for adapting large language models (LLMs), achieving competitive results across a wide range of benchmarks. Building on this idea, LoFIT\cite{yin2024lofit} identified task-specific critical attention points and trained offsets to modify hidden representations accordingly. IntervMerge\cite{osial2025parameter} further extend representation finetuning to model merging, employing task-specific interventions to alleviate representational bias. CRFT\cite{huang2025enhancing} apply representation finetuning to Chain-of-Thought reasoning, leveraging information flow analysis to identify and optimize key representations. Whereas representation finetuning has demonstrated competitive performance across various domains, existing approaches have primarily been applied to textual inputs. The potential of presentation finetuning in continual learning has not yet been explored, and further adaptation is required to make it suitable for downstream continual learning tasks.

\begin{figure*}[htb]
\centering
\includegraphics[width=0.85\textwidth]{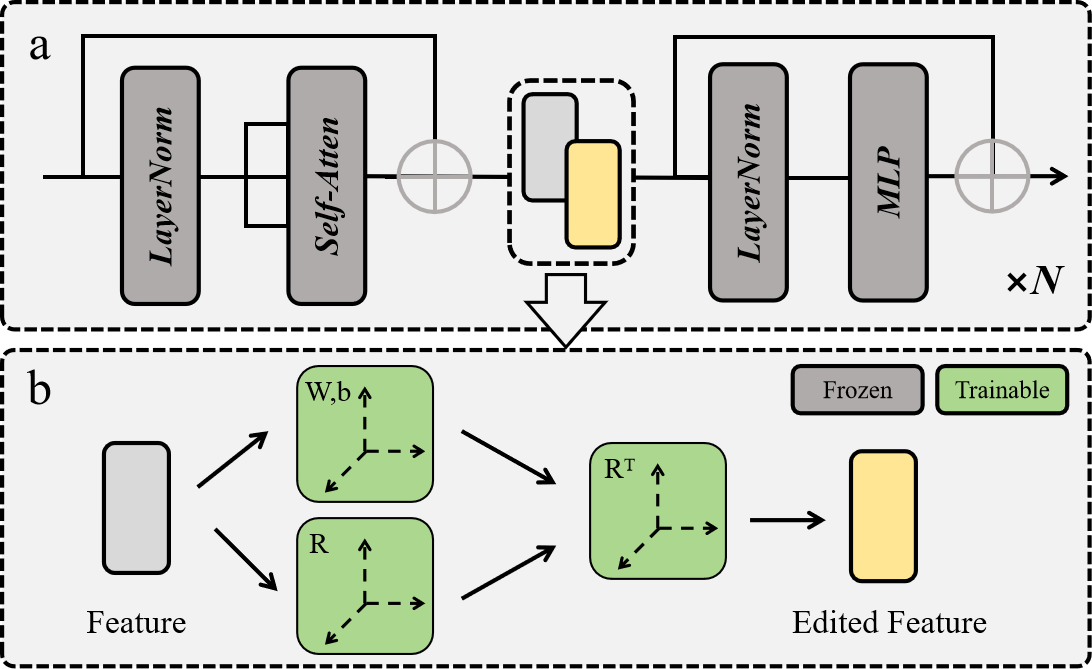}
\caption{The overall structure of the proposed method. (a) illustrates a standard ViT block and (b) depicts the implementation of ReFT. Unlike previous finetuning approaches, ReFT directly intervenes in the model by modifying its intermediate features. Specifically, the features are projected into a low-rank subspace via learnable parameters R, w, and b, and then mapped back to the original dimension. Gray areas indicate frozen parameters, and green areas denote trainable components.}
\label{fig:mian_framework}
\end{figure*}

\subsection{Parameter-Efficient tuning}
\label{subsec:peft}
The Parameter Efficient Fine-Tuning (PEFT) method finetunes the pre-trained model by incorporating lightweight modules. During the training process, the pretrained model remains frozen, and model intervention is done only by updating the lightweight modules. Representative methods include Adapters\cite{houlsby2019parameter}, Prompts\cite{jia2022visual}, and SSF\cite{lian2022scaling}. Adapter is a linear module, usually consisting of downsampling, upsampling and activation function. Prompt is a set of learnable vectors that are dynamically added to the hidden layer representation during computation. SSF adds scaling and displacement factors to the model weights, and learns the scaling and displacement factors to adjust the model.

\section{Method}
\label{sec:method}

\subsection{Continual Learning}
Given a sequence of tasks ${\mathcal{D}_1, \ldots, \mathcal{D}_T}$, here The $t$-th task is defined as $\mathcal{D}_t{=}\{(\bm{x}_i^t,\bm{y}_i^t)\}_{i=1}^{m_t}$,  where $\mathcal{D}_t$ contains $m_t$ samples $\bm{x}_i^t$ and their corresponding labels $\bm{y}_i^t$. Continual learning generally involves three scenarios: task-incremental learning(TIL)\cite{van2022three}, domain-incremental learning(DIL), and class-incremental learning(CIL). During the continual learning process, for instance in tasks $D_1$ and $D_2$, the input distributions generally differ, i.e.$P(X_1) \neq P(X_2)$, In TIL, both the target distributions $P(Y_1)$ and $P(Y_2)$, as well as the corresponding label spaces $\{Y_1\}$ and $\{Y_2\}$, are distinct. More importantly, the model has access to the task identifier $i$ (task id) during both training and inference. Consequently, at inference time, the model can select the corresponding classifier $\phi_i()$ based on the provided task id. This setting greatly reduces classification conflicts across tasks but depends on the availability of task identifiers, which limits its applicability when task information is unavailable. In DIL, although the data from different tasks originate from distinct domains, they share the same target label distribution, i.e., $ P (Y_1)=P(Y_2)$, and a common label space $\{Y_1\}=\{Y_2\}$. The categories remain consistent across domains, but the task identifier is not accessible during inference. This implies that the model cannot rely on task id for prediction, and must instead correctly classify data from unseen domains. CIL poses the most challenging scenario. In this case, the label distributions $P(Y_1) \neq P(Y_2)$ differ across tasks, and the label space $\{Y_1\} \subseteq \{Y_2\}$ continuously expands as new tasks introduce additional classes. During inference, the task id is not available and the model must dynamically adapt to the growing set of categories without relying on task information and retain the ability to recognize previously learned classes. This greatly increases the risk of catastrophic forgetting.

\subsection{Continual Learning with ReFT}
In large language models, taking transformer-based architectures as an example, the model first maps the input text into a semantic representation, which is then propagated through a sequence of block layers progressively learning the semantic representation. Based on the Domain Intervention Interpretation (DII) framework\cite{geiger2024finding}, given an initial hidden representation $\bm{h}_b$ of input text $b$ and representations $\bm{h}_s$ of counterfactual $s$, the distributed interchange intervention on $\bm{h}_b$ with the counterfactual source representation $\bm{h}_s$ could be defined as:
\begin{equation}
    \mathrm{DII}\left(\bm{h}_{b}, \bm{h}_{s}, R\right)=\bm{h}_b+R^{\top}\left(R \bm{h}_{s}-R \bm{h}_{b}\right)
\label{eq:DII}
\end{equation}
Where $R\in \mathbb{R}^{r \times d}$ is a low-rank projection matrix, $d$ is the semantic representation dimension, and $r$ is the rank of the subspace in which we intervene. (\ref{eq:DII}) provides a principle for counterfactual intervention in the model. \cite{wu2024reft} further proposed replacing the explicit counterfactual features with learnable linear matrix $W$ and $b$, introducing an orthogonal constraint that yields the representation editing formulation in large language models:
\begin{equation}
    \mathrm{\Phi}\left(\bm{h}_{b}, \bm{h}_{s}, R, W, b\right)=\bm{h}_b + R^{\top}\left(W\bm{h}_b+b-R \bm{h}_{b}\right)
\label{eq:reft}
\end{equation}

\begin{table*}[htb]
\centering
\caption{Performance comparison of different methods under the Task Incremental Learning scenario. All experiments are based on the ViT-B/16-IN21K. Each task consists of 5 classes for DTD and OxfordPet, 2 classes for EuroSAT and MNIST, 20 classes for StandCars and SUN397, and 10 classes for the remaining datasets. The best results are highlighted in bold.}
\renewcommand{\arraystretch}{1.5}
\small
\resizebox{1.0\textwidth}{!}{
\begin{tabular}{llccccccccccc}
\hline
& Method & 
 \rotatebox{90}{Aircraft} & 
 \rotatebox{90}{Clathch101} & 
 \rotatebox{90}{CIFAR} & 
 \rotatebox{90}{DTD} & 
 \rotatebox{90}{EuroSAT} & 
 \rotatebox{90}{Flowers} & 
 \rotatebox{90}{Food} & 
 \rotatebox{90}{MNIST} & 
 \rotatebox{90}{OxfordPet} & 
 \rotatebox{90}{StanfordCars } & 
 \rotatebox{90}{SUN397} \\
\hline
\multirow{6}{*}{Avg} & ViT\cite{dosovitskiy2020image} & 64.50 & 99.01 & 95.14 & 92.57 & 96.34 & 99.86 & 96.23 & 96.49 & 97.09 & 72.15 & 93.25 \\
 & Finetune & 67.94 & 99.36 & 95.03 & 92.39 & 89.60 & 99.81 & 94.12 & 96.38 & 97.01 & 70.66 & 93.11 \\
 & Prompt\cite{jia2022visual} & 69.33 & 99.29 & 98.01 & 90.04 & 89.84 & 99.79 & 96.61 & 94.60 & \textbf{97.56} & \textbf{79.92} & 93.44 \\
 & SSF\cite{lian2022scaling} & 66.63 & 99.31 & 97.52 & \textbf{93.85} & 94.39 & 99.81 & 95.96 & 97.36 & 96.98 & 75.11 & 93.14 \\
 & Adapter\cite{houlsby2019parameter} & 64.55 & 99.08 & 97.98 & 92.57 & 95.89 & 99.86 & 96.40 & 99.48 & 96.97 & 72.22 & 93.19 \\
 & CoRe(Ours) & \textbf{75.09} & \textbf{99.43} & \textbf{98.15} & 92.71 & \textbf{96.43} & \textbf{99.87} & \textbf{96.77} & \textbf{99.60} & 97.00 & 79.41 & \textbf{93.48} \\
\hline
\multirow{6}{*}{Last} & ViT\cite{dosovitskiy2020image} & 66.16 & 99.25 & 94.76 & 92.98 & 93.26 & \textbf{99.89} & 96.06 & 97.37 & \textbf{98.15} & 72.89 & 93.24 \\
 & Finetune & 66.07 & 99.20 & 94.61 & 93.19 & 85.33 & 99.84 & 94.28 & 96.31 & 97.33 & 69.63 & 93.11 \\
 & Prompt\cite{jia2022visual} & 69.37 & 99.03 & 97.45 & 90.96 & 86.31 & 99.82 & 96.34 & 94.93 & 97.98 & \textbf{80.74} & 93.26 \\
 & SSF\cite{lian2022scaling} & 71.05 & 99.41 & 96.90 & \textbf{94.31} & 90.24 & 99.85 & 95.67 & 97.43 & 97.89 & 75.46 & 93.03 \\
 & Adapter\cite{houlsby2019parameter} & 66.22 & 99.36 & 97.53 & 92.98 & 92.61 & \textbf{99.89} & 96.28 & 99.25 & 98.06 & 73.05 & 93.11 \\
 & CoRe(Ours) & \textbf{74.89} & \textbf{99.62} & \textbf{97.59} & 93.35 & \textbf{93.15} & \textbf{99.89} & \textbf{96.45} & \textbf{99.39} & 97.93 & 80.40 & \textbf{93.40} \\
\hline
\end{tabular}}
\label{tab:til}
\end{table*}

\begin{table*}[htb]
\centering
\caption{Performance comparison of different methods under the Domain Incremental Learning scenario. All experiments are conducted using ViT-B/16-IN21K. For all datasets, each task contains the same number of classes. The best results are highlighted in bold.}
\renewcommand{\arraystretch}{1.3}
\tiny
\resizebox{1.0\textwidth}{!}{
\begin{tabular}{lcccccccc}
\hline
 & \multicolumn{2}{c}{OfficeHome Inc65} & \multicolumn{2}{c}{CORe50 Inc50} & \multicolumn{2}{c}{CDDB Inc2} & \multicolumn{2}{c}{DomainNet Inc345} \\
Method & Avg & Last & Avg & Last & Avg & Last & Avg & Last  \\ \hline
ViT\cite{dosovitskiy2020image} & 74.22 & 64.99 & 77.55 & \textbf{61.49} & 76.30 & 72.41 & 50.73 & 40.37 \\
Finetune & 69.06 & 58.15 & 75.86 & 58.22 & 71.98 & 66.64 & 55.31 & 42.54 \\
Prompt\cite{jia2022visual} & 75.30 & 65.95 & 76.20 & 57.20 & 76.51 & 72.42 & 55.32 & 44.09 \\
SSF\cite{lian2022scaling} & 75.48 & 65.80 & 76.42 & 59.02 & 75.39 & 69.34 & 53.57 & 42.50 \\
Adapter\cite{houlsby2019parameter} & 74.67 & 65.83 & 76.50 & 58.53 & 76.64 & 72.07 & 55.51 & 44.23 \\
CoRe(Ours) & \textbf{75.96} & \textbf{66.30} & \textbf{77.95} & 59.91 & \textbf{77.35} & \textbf{72.50} & \textbf{56.73} & \textbf{45.23} \\ \hline
\end{tabular}}
\label{tab:dil}
\end{table*}

\begin{table*}[htb]
\centering
\caption{Performance comparison of various finetuning methods under Class Incremental Learning scenario. All experiments are based on ViT-B/16-IN21K. Here, 'IN-R' denotes ImageNet-R datasets, 'IN-A' refers to ImageNet-A datasets, 'ObjNet' represents the ObjectNet dataset, and 'Omni' stands for the OmniBenchmark dataset. For all datasets, each task contains an equal number of classes. The best results are highlighted in bold.}
\renewcommand{\arraystretch}{1.6}
\small
\resizebox{1.0\textwidth}{!}{
\begin{tabular}{w{l}{1.7cm}w{c}{0.5cm}w{c}{0.5cm}w{c}{0.4cm}w{c}{0.4cm}w{c}{0.4cm}w{c}{0.4cm}w{c}{0.5cm}w{c}{0.5cm}w{c}{0.4cm}w{c}{0.4cm}w{c}{0.4cm}w{c}{0.4cm}w{c}{0.4cm}w{c}{0.4cm}}
\hline
 & \multicolumn{2}{c}{CIFAR Inc5} & \multicolumn{2}{c}{CUB Inc10} & \multicolumn{2}{c}{IN-R Inc5} & \multicolumn{2}{c}{IN-A Inc10} & \multicolumn{2}{c}{ObjNet Inc10} & \multicolumn{2}{c}{Omni Inc30} & \multicolumn{2}{c}{VTAB Inc10} \\
Method & Avg & Last & Avg & Last & Avg & Last & Avg & Last & Avg & Last & Avg & Last & Avg & Last \\ \hline
ViT\cite{dosovitskiy2020image} & 87.57 & 81.26 & 92.20 & 86.73 & 62.58 & 54.55 & 60.5 & 49.44 & 65.45 & 53.59 & 79.34 & 73.15 & 85.99 & 84.38 \\
Finetune & 87.67 & 81.27 & 91.82 & 86.39 & 70.51 & 62.42 & 61.57 & 50.76 & 61.41 & 48.34 & 73.02 & 65.03 & 87.47 & 80.44 \\
Prompt\cite{jia2022visual} & 88.46 & 82.17 & 91.02 & 84.99 & 68.79 & 60.48 & 60.59 & 48.72 & 67.83 & 54.65 & 81.05 & 74.47 & 86.59 & 83.06 \\
SSF\cite{lian2022scaling} & 87.78 & 81.98 & 91.72 & 86.13 & 68.94 & 60.60 & \textbf{62.81} & \textbf{51.48} & 69.15 & 56.64 & 80.53 & 74.00 & 85.66 & 81.92 \\
Adapter\cite{houlsby2019parameter} & 90.65 & 85.15 & 92.21 & 86.73 & 72.35 & 64.33 & 60.53 & 49.57 & 67.18 & 55.24 & 80.75 & 74.37 & 85.95 & 84.35 \\
CoRe(Ours) & \textbf{91.32} & \textbf{86.04} & \textbf{92.51} & \textbf{86.90} & \textbf{72.52} & \textbf{64.40} & 61.64 & 49.05 & \textbf{71.00} & \textbf{58.59} & \textbf{81.50} & \textbf{75.04} & \textbf{89.42} & \textbf{85.65} \\ \hline
\end{tabular}}
\label{tab:cil}
\end{table*}

In the context of continual learning with visual inputs, representations requiring intervention can similarly be regarded as counterfactual information. For instance, if the model incorrectly classifies Samoyed's visual feature $\bm{e}_s$ as spotted dog $\bm{e}_b$. By applying (\ref{eq:reft}), the feature representation of a Samoyed $\bm{e}_s$ can be treated as counterfactual information to guide the model’s finetuning. As shown in Fig.\ref{fig:mian_framework}, for transformer-based pretrained models, such as ViT, the learned representations can similarly be modified through an learnable linear transformation: $W\bm{e}+b$ in place of explicit counterfactual. This leads to the following formulation, which removes the dependency on manually constructed counterfactual information:
\begin{equation}
    g_{\theta}(\bm{e}_b)=\bm{e}_b + R^{\top}\left(W\bm{e}_b+b-R \bm{e}_{b}\right)
\label{eq:visual_reft}
\end{equation}
Here, $R$ defines the intervention subspace, $W$ and $b$ are used to learn the calibration rule. The objective is to make $W\bm{e}_b+b$ approximate $\bm{e}_s$, thereby aligning the transformed representation $g_{\theta}(\bm{e}_b)$ more closely with the true feature $\bm{e}_s$. In this way, the original representation $\bm{e}_b$ is calibrated within a low-rank subspace, rather than being optimized in a black-box manner. With (\ref{eq:visual_reft}), the finetuning of pre-trained model $f_0$ could be described as:
\begin{equation}
f^{*}(\bm{x}) = \mathcal{F}\left(f_0(\bm{x}), \mathcal{D}_1,\theta\right)
\label{eq:finetune_vit}
\end{equation}
Where $f^{*}(\bm{x})$ is the finetuned model, $f_0$ is the pre-trained model, like ViT, $\theta$ is the trainable parameters of ReFT.

\subsection{Theoretical Foundation and Formulation}
Unlike traditional weight-based PEFT methods that operate through black-box optimization of model parameters, CoRe establishes a fundamentally different paradigm by performing explicit interventions in the representation space. The representation intervention introduced by CoRe admits the following norm bound.:

\textit{Representation Perturbation Bound}

Let the representation intervention $\Delta\bm{e}$ in the visual ReFT formulation be defined as:
\begin{equation}
    \Delta\bm{e} = R^{\top}(W\bm{e}+b-R\bm{e})
\end{equation}
By the sub-multiplicativity of matrix norms, we have:
\begin{align}
\|\Delta\bm{e}\|_2 
&= \|R^{\top}(W\bm{e}+b-R\bm{e})\|_2 \notag \\
&\leq \|R^{\top}\|_2 \cdot \|W\bm{e}+b-R\bm{e}\|_2 \notag \\
&\leq \sigma_{\max}(R^{\top}) \|W\bm{e}+b-R\bm{e}\|_2,
\end{align}
where $\sigma_{\max}(R^{\top})$ denotes the maximum singular value of $R^{\top}$. In particular, when enforcing the orthogonality constraint $R^{\top}R=I$, we have $\sigma_{\max}(R^{\top})=1$, yielding a tight upper bound on representation perturbations. Such bounded updates are expected to help mitigate catastrophic forgetting by limiting unintended representation drift across sequential tasks.

% The intervention constrained to a low-rank subspace $R$ satisfies:
% \begin{equation}
% \|\Delta\bm{e}\|_2 \leq \sigma_{\max}(R^{\top})\|(W-I)\bm{e}+b\|_2
% \end{equation}
% where $\sigma_{\max}(R^{\top})$ denotes the maximum singular value of $R^{\top}$. This result shows that the magnitude of representation change is explicitly bounded by the geometry of the intervention subspace.

% $Proof$

%\begin{proposition}[Explicit Optimization Objective]
\textit{Optimization Objective}

In CoRe, we provide an explicit optimization target that directly aligns calibrated representations with task-specific ideal representations. To learn the representation intervention for task $\mathcal{D}_t$, CoRe optimizes the following objective:
\begin{equation}
\min_{R_t, W_t, b_t}
\;\;
\|\bm{e}_s - g_{\theta_t}(\bm{e}_b)\|_2^2
\;+\; \|R_t^{\top}R_t-I\|_F^2,
\end{equation}
where $\bm{e}_s$ denotes the target representation, $\bm{e}_b$ is the base representation. The optimization objective consists of two components: the representation alignment term and the orthogonality constraint. The first term 
\begin{equation}
\|\bm{e}_s - g_{\theta_t}(\bm{e}_b)\|_2^2
\end{equation}
ensures that the transformed representation $g_{\theta_t}(\bm{e}_b)$ closely approximates the target representation $\bm{e}_s$. The second term \begin{equation}
\|R_t^{\top}R_t-I\|_F^2
\end{equation}
maintains the orthogonality of the projection matrix $R_t$, which is crucial for preserving the geometric properties of the subspace intervention. This combined objective provides a principled framework for representation calibration in continual learning.

%\end{proposition}

\section{Experiments}
\label{sec:exp}

\subsection{Implementation Details}
\label{exp:Implemen}

\noindent \textbf{Benchmark}
We evaluate our approach under three continual learning scenarios: task incremental learning, domain incremental learning, and class incremental learning. In the task incremental learning setting, following \cite{yu2024boosting}, we evaluate model performance on a collection of datasets, including Aircraft\cite{maji2013fine}, Caltech101\cite{fei2004learning}, CIFAR100\cite{krizhevsky2009learning}, DTD\cite{cimpoi2014describing}, EuroSAT\cite{helber2019eurosat}, Flowers102\cite{nilsback2008automated}, Food101\cite{bossard2014food}, MNIST\cite{lecun2002gradient}, OxfordPet\cite{parkhi2012cats}, StanfordCars\cite{krause20133d}, and SUN397\cite{xiao2010sun}. These datasets cover diverse characteristics: fine-grained datasets such as Aircraft and StanfordCars; broader-coverage datasets such as EuroSAT; and large-scale scene recognition datasets such as SUN397. For evaluation, EuroSAT and MNIST are divided into 5 tasks with 2 classes per task, and Oxford Pet is split into 8 tasks. In addition, All other datasets are partitioned into 10 tasks, each containing an equal number of classes. In the domain incremental learning setting, following \cite{wang2022s}, we conduct experiments on CDDB\cite{li2023continual}, CORe50\cite{lomonaco2017core50}, DomainNet\cite{peng2019moment}, and OfficeHome\cite{venkateswara2017deep}. Among them, CDDB focuses on user behavior recognition and consists of 10 domains; CORe50 contains 50 object classes captured under 11 different lighting and background conditions; DomainNet is the largest and most diverse benchmark for domain shift classification, consisting of 345 categories across 6 highly heterogeneous domains. In this setting, each domain is regarded as a separate task, and each task includes all categories. In the class incremental learning setting, following \cite{zhou2025revisiting}, we evaluate our method on CIFAR100\cite{krizhevsky2009learning}, CUB200\cite{wah2011caltech}, ImageNet-A\cite{hendrycks2021natural}, ImageNet-R\cite{hendrycks2021many}, ObjectNet\cite{barbu2019objectnet}, OmniBenchmark\cite{zerroug2022benchmark}, and VTAB\cite{zhai2019large}. These datasets include fine-grained recognition benchmarks such as CUB200, challenging subsets of ImageNet such as ImageNet-A and ImageNet-R, and large-scale benchmarks such as OmniBenchmark. In this setting, CIFAR100 and ImageNet-R are split into tasks with 5 classes per task, OmniBenchmark into tasks with 30 classes each, and the remaining datasets are divided into tasks containing 10 classes each. For clarity, we denote the number of classes in each task using the notation 'Inc'. For example, “Inc10” indicates that every task contains 10 classes.

\noindent \textbf{Baseline}
Model finetuning aims to adapt pre-trained models to downstream tasks by updating only a small fraction of parameters compared with the full model. We compare CoRe against several representative parameter-efficient fine-tuning methods, including Adapter\cite{houlsby2019parameter}, Prompt\cite{jia2022visual}, and SSF\cite{lian2022scaling}. Specifically, Adapter introduces trainable projection layers, typically consisting of down-sampling, up-sampling, and non-linear activation functions, into pretrained models. Prompt tuning employs a set of learnable vectors that are concatenated with hidden layer representations to modulate model outputs. SSF finetunes models by learning task-specific scaling and shifting parameters applied to model weights. In addition to these PEFT baselines, we also include comparisons with the frozen pretrained model outputs and full-finetuning.

\begin{figure}[htb]
\scriptsize
\centering
\includegraphics[width=0.48\textwidth]{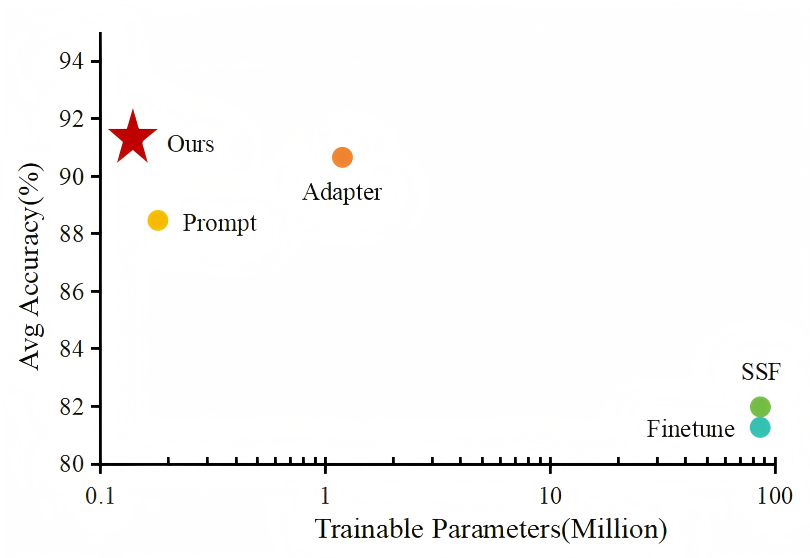}
%\vspace{-0.4cm}
\caption{Comparison of trainable parameters and Avg accuracy of finetuning methods, the experiments are conducted on CIFAR Inc10 with ViT-B/16-IN21K.}
\label{fig:para-avg}
\end{figure}

\noindent \textbf{Training Details}
We evaluate our method under different continual learning settings, including task incremental learning(TIL), domain incremental learning(DIL), and class incremental learning(CIL). During training, following \cite{zhou2025revisiting}, we train the model only on the first task and compute the class-wise feature means, which are then used as the classifier weights. For subsequent tasks, the model parameters remain frozen, and only the prototype-based classifier is updated. The classifier design differs across the three continual learning scenarios. In TIL, we construct a task-specific classifier for each task. During inference, the task identifier is provided to select the corresponding classifier for prediction. In DIL, following \cite{wang2022s}, we create a separate classifier and a set of k-means cluster centers for each domain during training. At inference time, the input is assigned to the nearest domain center, and the corresponding classifier is used. In CIL, the classifier is dynamically expanded as new classes arrive, and the model must predict over all previously seen classes without any external task information. All experiments are conducted with the ViT-B/16-IN21K and ViT-B/16-IN1K as backbone. The ViT-B/16-IN21K is pretrained on ImageNet21k and ViT-B/16-IN1K is further finetuned on ImageNet1k, which shifts the concerns of model. For finetuning, we adopt stochastic gradient descent (SGD) as the optimizer with an initial learning rate of 0.05, scheduled by cosine decay. A weight decay of 0.0005 is applied. The batch size is set to 48, and epochs is 20.

\noindent \textbf{Evaluation Metric}
In model performance evaluation, we follow \cite{zhou2025revisiting}, using Average accuracy (\textbf{Avg}) and Last accuracy (\textbf{Last}) to measure the performance of model. Specifically, \textbf{Last} denotes the Top-1 accuracy of every task, and \textbf{Avg} is the average value of \textbf{Last} of all tasks. Mathematically, for the $t$-th task, Average accuracy is calculated as follows: \textbf{Avg}${_t} = \frac{1}{t}\sum_{i=1}^{t}$\textbf{Last}$_i$.

\subsection{Main Results}
\label{exp:main_result}

We first evaluate CoRe against a range of representative finetuning baselines under the task-incremental learning (TIL) setting, and the results are summarized in Table.\ref{tab:til}. As shown in Table.\ref{tab:til}, CoRe achieves competitive performance across all datasets, demonstrating its ability to substantially enhance the feature extraction capability of pre-trained models in diverse scenarios. On fine-grained benchmarks such as Aircraft and Oxford Pets, CoRe effectively captures subtle intra-class distinctions. For datasets with pronounced domain shifts, such as EuroSAT, it adapts to distributional variations while maintaining training stability. Furthermore, its strong performance on large-scale datasets such as SUN397 underscores the scalability and robustness of CoRe in handling complex visual concepts. These results collectively validate CoRe as an effective and general solution for task-incremental learning.

We further assess CoRe under the domain-incremental learning (DIL) setting, where each task corresponds to a new domain while sharing an identical set of categories, requiring the model to capture domain-specific variations without sacrificing class discriminability. Following the experimental protocol of \cite{wang2022s}, we conduct evaluations on CDDB, CORe50, DomainNet, and OfficeHome, with the results summarized in Table.\ref{tab:dil}. As observed in Table.\ref{tab:dil}, CoRe consistently outperforms competing methods across all benchmarks, demonstrating its ability to learn domain-invariant representations while retaining sufficient flexibility to model domain-specific characteristics. This balanced representation learning allows CoRe to generalize effectively across shifts in appearance, style, and environmental context, while mitigating catastrophic forgetting. The strong and stable performance across multiple datasets highlights the robustness of CoRe in dynamic settings with continuously evolving data distributions.

\begin{figure}[htbp]
\scriptsize
\centering
    \includegraphics[width=0.48\textwidth]{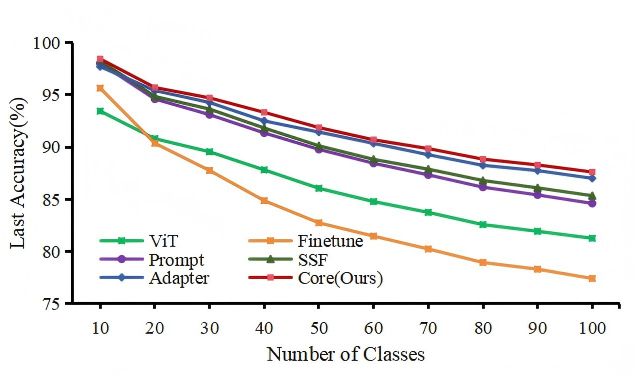}
    %\vspace{-0.4cm}
\caption{Average value of Last accuracy of methods under different random seeds. All the results are conducted on CIFAR Inc10 with ViT-B/16-IN21K.}
\label{fig:seed}
\end{figure}

\begin{table*}[htbp]
\centering
\caption{Performance comparison between CoRe and other finetune methods under different imbalance factors $\alpha$. All experiments are conducted on CIFAR Inc10 with ViT-B/16-IN21K, the best results are highlighted in bold.}
\label{tab:imbalance}
\renewcommand{\arraystretch}{1.1}
\resizebox{1.0\textwidth}{!}{
\begin{tabular}{lcccccccccc}
\toprule
 & \multicolumn{2}{c}{$\alpha=$ 1} & \multicolumn{2}{c}{$\alpha=$ 0.5} & \multicolumn{2}{c}{$\alpha=$ 0.1} & \multicolumn{2}{c}{$\alpha=$ 0.05} & \multicolumn{2}{c}{$\alpha=$ 0.01} \\
Method & Avg & Last & Avg & Last & Avg & Last & Avg & Last & Avg & Last \\
\midrule
ViT\cite{dosovitskiy2020image} & 87.13 & 81.26 & 87.07 & 81.14 & 86.70 & 80.88 & 86.51 & 80.47 & 84.70 & 78.07 \\
Finetune & 88.56 & 82.80 & 88.06 & 82.52 & 88.40 & 83.11 & 88.06 & 82.24 & 85.70 & 79.68 \\
Prompt\cite{jia2022visual} & 90.59 & 85.27 & 90.51 & 85.12 & 88.43 & 84.99 & 86.83 & 80.45 & 76.61 & 68.52 \\
SSF\cite{lian2022scaling} & 90.71 & 85.21 & 90.55 & 85.09 & 90.17 & 84.88 & 89.71 & 84.38 & 87.03 & 80.96 \\
Adapter\cite{houlsby2019parameter} & 92.27 & 87.50 & 91.94 & 87.14 & 89.51 & 84.15 & 88.29 & 82.69 & 85.25 & 79.09 \\
CoRe(Ours) & \textbf{92.37} & \textbf{87.60} & \textbf{92.34} & \textbf{87.58} & \textbf{91.76} & \textbf{86.95} & \textbf{91.87} & \textbf{86.89} & \textbf{89.41} & \textbf{83.23} \\
\bottomrule
\end{tabular}}
\end{table*}

Finally, we assess all methods in the class incremental learning setting, which is widely regarded as the most challenging scenario in continual learning. In this setting, the task id is unknown during inference, and the model must classify samples among all previously seen classes. This makes the problem substantially harder but also more reflective of real-world applications. Following \cite{zhou2025revisiting}, we evaluate the methods on multiple benchmarks, and the results are summarized in Table.\ref{tab:cil}. From Table.\ref{tab:cil}, we can observe that CoRe outperforms other finetuning methods in most cases. These results confirm that CoRe effectively retains knowledge from previously learned tasks and acquires new representations in highly challenging settings. By operating in the representation subspace rather than directly modifying model weights, CoRe mitigates catastrophic forgetting and preserves discriminative features of past classes. This ability to balance stability and plasticity makes CoRe particularly well-suited for real-world continual learning scenarios where task identity is not available at inference.

\subsection{Model Analysis}
\label{exp:papa_efficien}

In the context of model finetuning, parameter efficiency is a critical factor, particularly in continual learning settings where models are repeatedly updated as new tasks arrive. We compare CoRe with several representative finetuning methods in terms of both the number of trainable parameters and Avg accuracy. All experiments are conducted on CIFAR Inc10 under the class-incremental learning protocol, using ViT-B/16-IN21k as the backbone and the results are summarized in Fig.\ref{fig:para-avg}. As shown in Fig.\ref{fig:para-avg}, CoRe achieves the highest average accuracy while requiring the fewest trainable parameters among all compared approaches. By updating only task-relevant subspaces and preserving the majority of pretrained representations, CoRe effectively balances performance and efficiency. This design enables effective continual adaptation with low computational overhead, highlighting the value of representation-level interventions for scalable and efficient continual learning.

During the training phase of CoRe, we fix the random seed to 1993 when shuffling the order of all datasets, ensuring a deterministic and reproducible training sequence. This controlled setup avoids potential biases introduced by a specific data ordering and allows for a fair comparison across different methods under identical conditions. To further assess the robustness of CoRe and rule out performance gains arising from favorable random initialization or task ordering, we conduct additional experiments using multiple random seeds \{1991, 1992, 1993, 1994, 1995\}. For each seed, both the data order and model initialization are varied, and we report the averaged Last accuracy across all runs in Fig.\ref{fig:seed}. As observed in Fig.\ref{fig:seed}, CoRe consistently outperforms competing approaches across all random seeds. These results demonstrate that CoRe not only achieves superior accuracy but also maintains stable performance under different stochastic conditions. This stability indicates that the effectiveness of CoRe is intrinsic to its design rather than dependent on fortuitous random choices during training.

\begin{figure*}[htb]
\scriptsize
\centering
    \begin{subfigure}{0.33 \textwidth}
        \includegraphics[width=\textwidth]{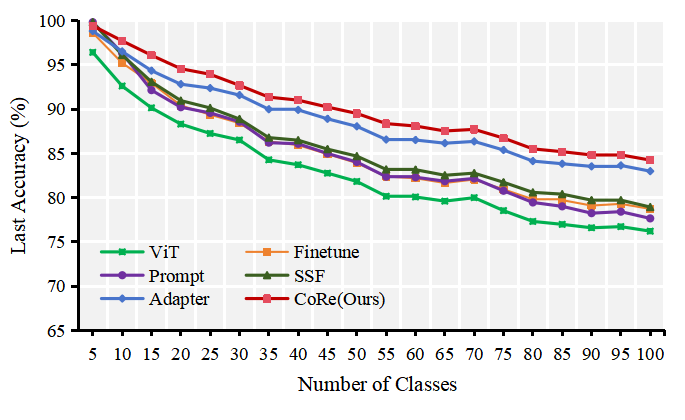}
        \vspace{-0.4cm}
        \caption{CIFAR Inc5}
    \end{subfigure}
    \begin{subfigure}{0.33\textwidth}
        \includegraphics[width=\textwidth]{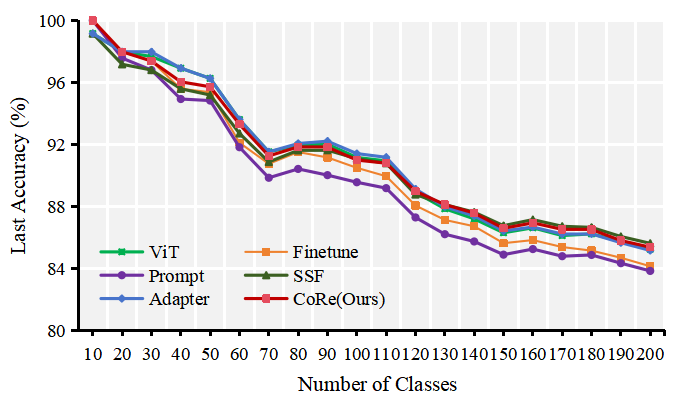}
        \vspace{-0.4cm}
        \caption{CUB Inc10}
    \end{subfigure}
    %\%vspace{0.02\textwidth}
    \begin{subfigure}{0.33\textwidth}
        \includegraphics[width=\textwidth]{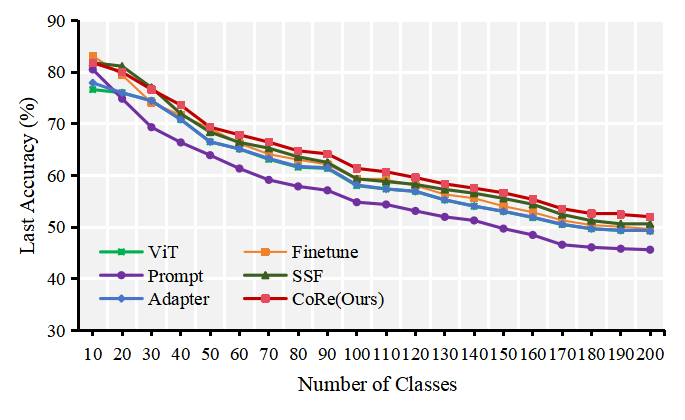}
        \vspace{-0.4cm}
        \caption{ImageNet-A Inc10}
    \end{subfigure}
    \\
    \begin{subfigure}{0.33\textwidth}
        \includegraphics[width=\textwidth]{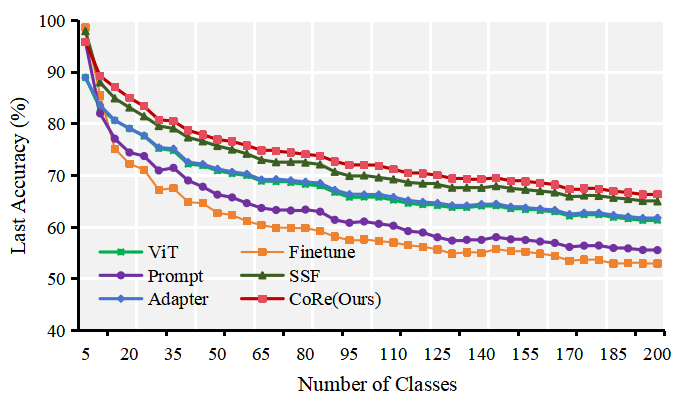}
        \vspace{-0.4cm}
        \caption{ImageNet-R Inc5}
    \end{subfigure} 
        \begin{subfigure}{0.33\textwidth}
        \includegraphics[width=\textwidth]{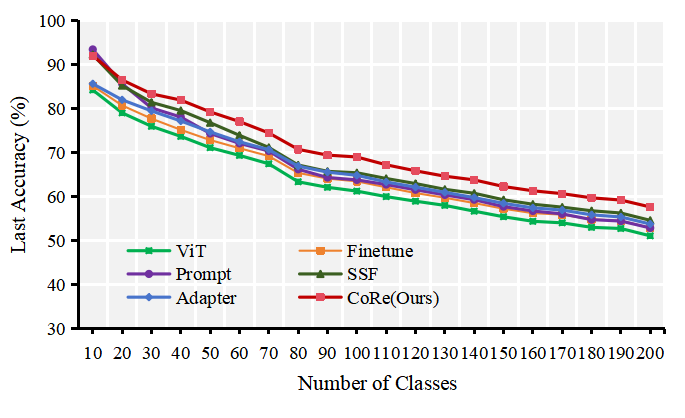}
        \vspace{-0.4cm}
        \caption{ObjectNet Inc10}
    \end{subfigure}
    %\hspace{0.01\textwidth}
    \begin{subfigure}{0.33\textwidth}
        \includegraphics[width=\textwidth]{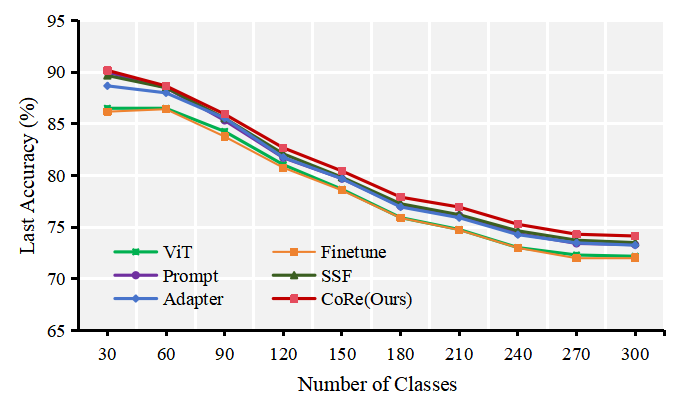}
        \vspace{-0.4cm}
        \caption{Omnibenchmark Inc30}
    \end{subfigure}
    \caption{Last accuracy of various finetuning methods under Class Incremental Learning scenario using ViT-B/16-IN1K as the backbone. Each task contains the same number of classes across all datasets. Core consistently outperforms other finetuning approaches even with the ViT-B/16-IN1K architecture.}
    \label{fig:per_task}
\end{figure*}

In real-world scenarios, task data are rarely uniformly distributed, and continual learning systems must often contend with varying degrees of class imbalance across tasks. Such imbalance can significantly bias representation learning, causing models to overfit majority classes while under-representing minority ones. To investigate the robustness of different methods under this challenging yet practical setting, we evaluate the performance of different methods under class-imbalanced settings. The experiments are conducted on CIFAR Inc10 with ViT-B/16-IN21K as backbone, and results are summarized in Table.\ref{tab:imbalance}, where $\alpha$ denotes the imbalance ratio. Specifically, the number of samples for class $i$ is computed as $M * (\alpha^{(i/N})$ where $M$ is the original number of samples per class in the balanced setting, and $N$ is the total number of classes. As shown in Table.\ref{tab:imbalance}, model performance gradually declines as the imbalance ratio decreases, which highlights that class imbalance poses a substantial challenge for continual learning and lead to biased feature learning and degraded generalization. Nonetheless, CoRe consistently outperforms other methods across varying imbalance ratios, demonstrating its robustness and stability under imbalanced conditions.

In addition, we further evaluate model performance using ViT-B/16-IN1k as the backbone. Unlike ViT-B/16-IN21k, this model is first pretrained on ImageNet-21k and then finetuned on ImageNet-1k, which leads to a different feature bias and optimization focus. To examine whether CoRe remains effective under such a change in pretraining history, we conduct experiments on CIFAR Inc10 in the class incremental learning setting and compare the Last accuracy across different methods, and the results are reported in Fig.\ref{fig:per_task}. As we can see from Fig.\ref{fig:per_task}, CoRe consistently achieves superior performance in most cases. This indicates that the advantage of CoRe is not tied to a specific pretrained backbone, but generalizes well across models with different pretraining and finetuning pipelines, demonstrating its robustness under diverse backbone initializations.

\begin{table}[tbp]
\centering
\normalsize %\tiny \scriptsize \footnotesize \small \normalsize \large \Large
\caption{Ablation studies on ReFT rank, where "Reft Rank" indicates the rank value of every ReFT in model. All experiments are conducted on CIFAR Inc10 with ViT-B/16-IN21K.}

\renewcommand{\arraystretch}{1.2}
\begin{tabular}{p{2.5cm}p{2.4cm}p{2.4cm}}
  \toprule
  Reft Rank & Avg & Last\\
  \midrule
  4  & 92.28  & 87.62\\
  8  & 92.34  & 87.63\\
  16  & 92.27  & 87.51\\
  32 & 92.02  & 87.29\\
  64 & 92.01 & 87.08 \\
  \bottomrule
\end{tabular}%}
\label{tab:abala_rank}
\end{table}

\begin{table}[tbp]
\centering
\normalsize %\scriptsize \footnotesize \small \normalsize \large \Large
\caption{Ablation studies on the number of block layer inserted with ReFT. The ViT-B/16-IN21K architecture consists of 12 block layers. Here, "1" indicates that ReFT is inserted into only one block layer, and "12" denotes that ReFT is applied to every block layer. All experiments are conducted on CIFAR Inc10 with ViT-B/16-IN21K.}
\renewcommand{\arraystretch}{1.2}
\begin{tabular}{p{2.5cm}p{2.4cm}p{2.4cm}}
\toprule
  Layer & Avg & Last\\
  \midrule
 1  & 91.95  & 87.25 \\
 3  & 91.59  & 86.61 \\
 6  & 92.16  & 87.40 \\
 9  & 92.17  & 87.59 \\
 12  & 92.24  & 87.63 \\
 \bottomrule
\end{tabular}%}
\label{tab:abala_layer}
\end{table}

\subsection{Ablation Study}
\label{exp:ablation}

In the CoRe method, feature representations are updated within a low-rank subspace, where the rank of the subspace serves as a pivotal hyperparameter that critically influences continual learning performance. To investigate the impact of rank selection, we conduct experiments on the CIFAR Inc10 under the class-incremental learning setting using ViT-B/16-IN21K as backbone, and the results are summarized in Table.\ref{tab:abala_rank}. As we can see from Table.\ref{tab:abala_rank}, model performance improves substantially when the rank increases from a small initial value, indicating that a larger subspace provides the necessary capacity to capture and preserve task-specific feature variations, thereby mitigating catastrophic forgetting. However, performance begins to degrade when the rank exceeds a certain threshold, which may caused by increased parameter redundancy and can lead to overfitting on recent tasks. Furthermore, a higher rank directly corresponds to greater computational overhead, both in terms of memory and training time. Considering this performance-efficiency trade-off, we select a rank of 8 for our method.

The ViT architecture is composed of a sequence of transformer blocks, and finetuning different numbers of layers affects the degree of model intervention. We systematically examine the effect of inserting ReFT into an incrementally increasing number of blocks on the CIFAR Inc10 benchmark with ViT-B/16-IN21K as backbone, and detailed results are presented in Table.\ref{tab:abala_layer}. As shown in Table.\ref{tab:abala_layer}, applying ReFT to more block layers enhances model accuracy, which allows for deeper, more pervasive model adaptation, and the best overall performance is achieved when all 12 blocks are equipped. However, the relationship is not strictly monotonic. The configuration with ReFT in only a single block occasionally outperforms the configuration with three blocks, suggesting that the placement of adaptation modules interacts in a nonlinear way with the model's performance. Based on these observations, we insert ReFT into all 12 block layers to fully leverage the hierarchical feature representations.

\section{Conclusion}
\label{sec:conclusion}
In this work, we introduced CoRe, the first framework that integrates representation finetuning into continual learning. Unlike conventional parameter-efficient finetuning approaches that intervene primarily at the weight level, CoRe performs task-specific interventions in a low-rank subspace of hidden representations and adopts a learning paradigm with explicit objectives rather than relying on black-box optimization, enabling efficient adaptation to new tasks and mitigating catastrophic forgetting. Through extensive experiments, we demonstrated that CoRe consistently outperforms existing finetuning baselines, achieving state-of-the-art performance and maintaining parameter efficiency. These results highlight the potential of representation-level interventions as an effective and scalable alternative to weight-based adaptation in dynamic learning environments. Our work extends the applicability of ReFT to continual learning for the first time, providing insights for future research on representation-level adaptation in large pre-trained models.

% \section*{Acknowledgment}

% The preferred spelling of the word ``acknowledgment'' in American English is without an ``e'' after the ``g.'' Use the singular heading even if you have many acknowledgments. Avoid expressions such as ``One of us (S.B.A.) would like to thank ... .'' Instead, write ``F. A. Author thanks ... .'' In most cases, sponsor and financial support acknowledgments are placed in the unnumbered footnote on the first page, not here.

%\newpage


\subsection*{References}

\begin{thebibliography}{00}

\bibitem{goodfellow2013empirical}I. J. Goodfellow, M. Mirza, D. Xiao, A. Courville, and Y. Bengio, ``An empirical investigation of catastrophic forgetting in gradient-based neural networks,'' {\em arXiv preprint arXiv:1312.6211}, 2013.

\bibitem{de2021continual}M. De Lange, R. Aljundi, M. Masana, S. Parisot, X. Jia, A. Leonardis, G. Slabaugh, and T. Tuytelaars, ``A continual learning survey: Defying forgetting in classification tasks,'' {\em IEEE Trans. Pattern Anal. Mach. Intell.}, vol. 44, no. 7, pp. 3366--3385, Jul. 2021.

\bibitem{masana2022class}M. Masana, X. Liu, B. Twardowski, M. Menta, A. D. Bagdanov, and J. Van De Weijer, ``Class-incremental learning: survey and performance evaluation on image classification,'' {\em IEEE Trans. Pattern Anal. Mach. Intell.}, vol. 45, no. 5, pp. 5513--5533, May 2022.

\bibitem{wang2022learning}Z. Wang, Z. Zhang, C.-Y. Lee, H. Zhang, R. Sun, X. Ren, G. Su, V. Perot, J. Dy, and T. Pfister, ``Learning to prompt for continual learning,'' in {\em Proc. IEEE/CVF Conf. Comput. Vis. Pattern Recognit.}, 2022, pp. 139--149.

\bibitem{wang2022dualprompt}Z. Wang, Z. Zhang, S. Ebrahimi, R. Sun, H. Zhang, C.-Y. Lee, X. Ren, G. Su, V. Perot, J. Dy, and others, ``Dualprompt: Complementary prompting for rehearsal-free continual learning,'' in {\em Eur. Conf. Comput. Vis.}, 2022, pp. 631--648.

\bibitem{yu2024boosting}J. Yu, Y. Zhuge, L. Zhang, P. Hu, D. Wang, H. Lu, and Y. He, ``Boosting continual learning of vision-language models via mixture-of-experts adapters,'' in {\em Proc. IEEE/CVF Conf. Comput. Vis. Pattern Recognit.}, 2024, pp. 23219--23230.

\bibitem{dosovitskiy2020image}A. Dosovitskiy, L. Beyer, A. Kolesnikov, D. Weissenborn, X. Zhai, T. Unterthiner, M. Dehghani, M. Minderer, G. Heigold, S. Gelly, and others, ``An image is worth 16x16 words: Transformers for image recognition at scale,'' {\em arXiv preprint arXiv:2010.11929}, 2020.

\bibitem{zhou2025revisiting}D.-W. Zhou, Z.-W. Cai, H.-J. Ye, D.-C. Zhan, and Z. Liu, ``Revisiting class-incremental learning with pre-trained models: Generalizability and adaptivity are all you need,'' {\em Int. J. Comput. Vis.}, vol. 133, no. 3, pp. 1012--1032, 2025.

\bibitem{houlsby2019parameter}N. Houlsby, A. Giurgiu, S. Jastrzebski, B. Morrone, Q. De Laroussilhe, A. Gesmundo, M. Attariyan, and S. Gelly, ``Parameter-efficient transfer learning for NLP,'' in {\em Int. Conf. Mach. Learn.}, 2019, pp. 2790--2799.

\bibitem{jia2022visual}M. Jia, L. Tang, B.-C. Chen, C. Cardie, S. Belongie, B. Hariharan, and S.-N. Lim, ``Visual prompt tuning,'' in {\em Eur. Conf. Comput. Vis.}, 2022, pp. 709--727.

\bibitem{wu2024reft}Z. Wu, A. Arora, Z. Wang, A. Geiger, D. Jurafsky, C. D. Manning, and C. Potts, ``Reft: Representation finetuning for language models,'' {\em Adv. Neural Inf. Process. Syst.}, vol. 37, pp. 63908--63962, 2024.

\bibitem{liu2025unlocking}T. Liu, R. Li, Y. Qi, H. Liu, X. Tang, T. Zheng, Q. Yin, M. X. Cheng, J. Huan, H. Wang, and others, ``Unlocking efficient, scalable, and continual knowledge editing with basis-level representation fine-tuning,'' {\em arXiv preprint arXiv:2503.00306}, 2025.

\bibitem{yin2024lofit}F. Yin, X. Ye, and G. Durrett, ``Lofit: Localized fine-tuning on llm representations,'' {\em Adv. Neural Inf. Process. Syst.}, vol. 37, pp. 9474--9506, 2024.

\bibitem{osial2025parameter}M. Osial, D. Marczak, and B. Zieli{\'n}ski, ``Parameter-efficient interventions for enhanced model merging,'' in {\em Proc. SIAM Int. Conf. Data Mining}, 2025, pp. 516--526.

\bibitem{huang2025enhancing}C. Huang, S. Yan, L. Xie, B. Lin, S. Fan, Y. Xin, D. Cai, C. Shen, and J. Ye, ``Enhancing Chain-of-Thought Reasoning with Critical Representation Fine-tuning,'' {\em arXiv preprint arXiv:2507.10085}, 2025.

\bibitem{aljundi2018memory}R. Aljundi, F. Babiloni, M. Elhoseiny, M. Rohrbach, and T. Tuytelaars, ``Memory aware synapses: Learning what (not) to forget,'' in {\em Proc. Eur. Conf. Comput. Vis.}, 2018, pp. 139--154.

\bibitem{kirkpatrick2017overcoming}J. Kirkpatrick, R. Pascanu, N. Rabinowitz, J. Veness, G. Desjardins, A. A. Rusu, K. Milan, J. Quan, T. Ramalho, A. Grabska-Barwinska, and others, ``Overcoming catastrophic forgetting in neural networks,'' {\em Proc. Natl. Acad. Sci.}, vol. 114, no. 13, pp. 3521--3526, Mar. 2017.

\bibitem{li2017learning}Z. Li and D. Hoiem, ``Learning without forgetting,'' {\em IEEE Trans. Pattern Anal. Mach. Intell.}, vol. 40, no. 12, pp. 2935--2947, Dec. 2017.

\bibitem{mallya2018packnet}A. Mallya and S. Lazebnik, ``Packnet: Adding multiple tasks to a single network by iterative pruning,'' in {\em Proc. IEEE Conf. Comput. Vis. Pattern Recognit.}, 2018, pp. 7765--7773.

\bibitem{serra2018overcoming}J. Serra, D. Suris, M. Miron, and A. Karatzoglou, ``Overcoming catastrophic forgetting with hard attention to the task,'' in {\em Int. Conf. Mach. Learn.}, 2018, pp. 4548--4557.

\bibitem{wang2020learn}Z. Wang, T. Jian, K. Chowdhury, Y. Wang, J. Dy, and S. Ioannidis, ``Learn-prune-share for lifelong learning,'' in {\em Proc. IEEE Int. Conf. Data Mining}, 2020, pp. 641--650.

\bibitem{rebuffi2017icarl}S.-A. Rebuffi, A. Kolesnikov, G. Sperl, and C. H. Lampert, ``icarl: Incremental classifier and representation learning,'' in {\em Proc. IEEE Conf. Comput. Vis. Pattern Recognit.}, 2017, pp. 2001--2010.

\bibitem{buzzega2020dark}P. Buzzega, M. Boschini, A. Porrello, D. Abati, and S. Calderara, ``Dark experience for general continual learning: a strong, simple baseline,'' {\em Adv. Neural Inf. Process. Syst.}, vol. 33, pp. 15920--15930, 2020.

\bibitem{cha2021co2l}H. Cha, J. Lee, and J. Shin, ``Co2l: Contrastive continual learning,'' in {\em Proc. IEEE/CVF Int. Conf. Comput. Vis.}, 2021, pp. 9516--9525.

\bibitem{radford2021learning}A. Radford, J. W. Kim, C. Hallacy, A. Ramesh, G. Goh, S. Agarwal, G. Sastry, A. Askell, P. Mishkin, J. Clark, and others, ``Learning transferable visual models from natural language supervision,'' in {\em Int. Conf. Mach. Learn.}, 2021, pp. 8748--8763.

\bibitem{lian2022scaling}D. Lian, D. Zhou, J. Feng, and X. Wang, ``Scaling \& shifting your features: A new baseline for efficient model tuning,'' {\em Adv. Neural Inf. Process. Syst.}, vol. 35, pp. 109--123, 2022.

\bibitem{van2022three}G. M. Van de Ven, T. Tuytelaars, and A. S. Tolias, ``Three types of incremental learning,'' {\em Nat. Mach. Intell.}, vol. 4, no. 12, pp. 1185--1197, Dec. 2022.

\bibitem{geiger2024finding}A. Geiger, Z. Wu, C. Potts, T. Icard, and N. Goodman, ``Finding alignments between interpretable causal variables and distributed neural representations,'' in {\em Causal Learn. Reasoning}, 2024, pp. 160--187.

\bibitem{wang2022s}Y. Wang, Z. Huang, and X. Hong, ``S-prompts learning with pre-trained transformers: An occam's razor for domain incremental learning,'' {\em Adv. Neural Inf. Process. Syst.}, vol. 35, pp. 5682--5695, 2022.

\bibitem{maji2013fine}S. Maji, E. Rahtu, J. Kannala, M. Blaschko, and A. Vedaldi, ``Fine-grained visual classification of aircraft,'' {\em arXiv preprint arXiv:1306.5151}, 2013.

\bibitem{fei2004learning}L. Fei-Fei, R. Fergus, and P. Perona, ``Learning generative visual models from few training examples: An incremental bayesian approach tested on 101 object categories,'' in {\em IEEE Conf. Comput. Vis. Pattern Recognit. Workshop}, 2004, pp. 178--178.

\bibitem{krizhevsky2009learning}A. Krizhevsky, G. Hinton, and others, ``Learning multiple layers of features from tiny images,'' 2009.

\bibitem{cimpoi2014describing}M. Cimpoi, S. Maji, I. Kokkinos, S. Mohamed, and A. Vedaldi, ``Describing textures in the wild,'' in {\em Proc. IEEE Conf. Comput. Vis. Pattern Recognit.}, 2014, pp. 3606--3613.

\bibitem{helber2019eurosat}P. Helber, B. Bischke, A. Dengel, and D. Borth, ``Eurosat: A novel dataset and deep learning benchmark for land use and land cover classification,'' {\em IEEE J. Sel. Top. Appl. Earth Observ. Remote Sens.}, vol. 12, no. 7, pp. 2217--2226, Jul. 2019.

\bibitem{nilsback2008automated}M.-E. Nilsback and A. Zisserman, ``Automated flower classification over a large number of classes,'' in {\em Indian Conf. Comput. Vis., Graph. Image Process.}, 2008, pp. 722--729.

\bibitem{bossard2014food}L. Bossard, M. Guillaumin, and L. Van Gool, ``Food-101--mining discriminative components with random forests,'' in {\em Eur. Conf. Comput. Vis.}, 2014, pp. 446--461.

\bibitem{lecun2002gradient}Y. LeCun, L. Bottou, Y. Bengio, and P. Haffner, ``Gradient-based learning applied to document recognition,'' {\em Proc. IEEE}, vol. 86, no. 11, pp. 2278--2324, Nov. 2002.

\bibitem{parkhi2012cats}O. M. Parkhi, A. Vedaldi, A. Zisserman, and C. V. Jawahar, ``Cats and dogs,'' in {\em IEEE Conf. Comput. Vis. Pattern Recognit.}, 2012, pp. 3498--3505.

\bibitem{krause20133d}J. Krause, M. Stark, J. Deng, and L. Fei-Fei, ``3d object representations for fine-grained categorization,'' in {\em Proc. IEEE Int. Conf. Comput. Vis. Workshops}, 2013, pp. 554--561.

\bibitem{xiao2010sun}J. Xiao, J. Hays, K. A. Ehinger, A. Oliva, and A. Torralba, ``Sun database: Large-scale scene recognition from abbey to zoo,'' in {\em IEEE Conf. Comput. Vis. Pattern Recognit.}, 2010, pp. 3485--3492.

\bibitem{li2023continual}C. Li, Z. Huang, D. P. Paudel, Y. Wang, M. Shahbazi, X. Hong, and L. Van Gool, ``A continual deepfake detection benchmark: Dataset, methods, and essentials,'' in {\em Proc. IEEE/CVF Winter Conf. Appl. Comput. Vis.}, 2023, pp. 1339--1349.

\bibitem{lomonaco2017core50}V. Lomonaco and D. Maltoni, ``Core50: a new dataset and benchmark for continuous object recognition,'' in {\em Conf. Robot Learn.}, 2017, pp. 17--26.

\bibitem{peng2019moment}X. Peng, Q. Bai, X. Xia, Z. Huang, K. Saenko, and B. Wang, ``Moment matching for multi-source domain adaptation,'' in {\em Proc. IEEE/CVF Int. Conf. Comput. Vis.}, 2019, pp. 1406--1415.

\bibitem{venkateswara2017deep}H. Venkateswara, J. Eusebio, S. Chakraborty, and S. Panchanathan, ``Deep hashing network for unsupervised domain adaptation,'' in {\em Proc. IEEE Conf. Comput. Vis. Pattern Recognit.}, 2017, pp. 5018--5027.

\bibitem{wah2011caltech}C. Wah, S. Branson, P. Welinder, P. Perona, and S. Belongie, ``The caltech-ucsd birds-200-2011 dataset,'' 2011.

\bibitem{hendrycks2021natural}D. Hendrycks, K. Zhao, S. Basart, J. Steinhardt, and D. Song, ``Natural adversarial examples,'' in {\em Proc. IEEE/CVF Conf. Comput. Vis. Pattern Recognit.}, 2021, pp. 15262--15271.

\bibitem{hendrycks2021many}D. Hendrycks, S. Basart, N. Mu, S. Kadavath, F. Wang, E. Dorundo, R. Desai, T. Zhu, S. Parajuli, M. Guo, and others, ``The many faces of robustness: A critical analysis of out-of-distribution generalization,'' in {\em Proc. IEEE/CVF Int. Conf. Comput. Vis.}, 2021, pp. 8340--8349.

\bibitem{barbu2019objectnet}A. Barbu, D. Mayo, J. Alverio, W. Luo, C. Wang, D. Gutfreund, J. Tenenbaum, and B. Katz, ``Objectnet: A large-scale bias-controlled dataset for pushing the limits of object recognition models,'' {\em Adv. Neural Inf. Process. Syst.}, vol. 32, 2019.

\bibitem{zerroug2022benchmark}A. Zerroug, M. Vaishnav, J. Colin, S. Musslick, and T. Serre, ``A benchmark for compositional visual reasoning,'' {\em Adv. Neural Inf. Process. Syst.}, vol. 35, pp. 29776--29788, 2022.

\bibitem{zhai2019large}X. Zhai, J. Puigcerver, A. Kolesnikov, P. Ruyssen, C. Riquelme, M. Lucic, J. Djolonga, A. S. Pinto, M. Neumann, A. Dosovitskiy, and others, ``A large-scale study of representation learning with the visual task adaptation benchmark,'' {\em arXiv preprint arXiv:1910.04867}, 2019.

\end{thebibliography}
\end{document}